\documentclass{article}

   \usepackage[nonatbib,preprint]{neurips_2023}

\usepackage[utf8]{inputenc} % allow utf-8 input
\usepackage[T1]{fontenc}    % use 8-bit T1 fonts
\usepackage{hyperref}       % hyperlinks
\usepackage{url}            % simple URL typesetting
\usepackage{booktabs}       % professional-quality tables
\usepackage{amsfonts}       % blackboard math symbols
\usepackage{nicefrac}       % compact symbols for 1/2, etc.
\usepackage{microtype}      % microtypography
\usepackage{xcolor}         % colors

\usepackage{caption}
\usepackage{subcaption}
\usepackage{graphicx}

\title{Dual policy as self-model for planning}

\author{%
  Jaesung Yoo\\
  Department of Brain and Cognitive Sciences\\
  Massachusetts Institute of Technology\\
  Cambridge, MA, 02139 \\
  \texttt{jsyoo61@mit.edu} \\
  \And
  Fernanda de la Torre \\
  Department of Brain and Cognitive Sciences\\
  Massachusetts Institute of Technology\\
  Cambridge, MA, 02139 \\
  \texttt{dlatorre@mit.edu}\\
  \And
  Guangyu Robert Yang \\
  Department of Brain and Cognitive Sciences\\
  Massachusetts Institute of Technology\\
  Cambridge, MA, 02139 \\
  \texttt{yanggr@mit.edu}\\
}

\begin{document}

\maketitle

\begin{abstract}
Planning is a data efficient decision-making strategy where an agent selects candidate actions by exploring possible future states. To simulate future states when there is a high-dimensional action space, the knowledge of one's decision making strategy must be used to limit the number of actions to be explored. We refer to the model used to simulate one's decisions as the agent's \textit{self-model}. While self-models are implicitly used widely in conjunction with world models to plan actions, it remains unclear how self-models should be designed. Inspired by current reinforcement learning approaches and neuroscience, we explore the benefits and limitations of using a distilled policy network as the self-model. In such \textit{dual-policy} agents, a model-free policy and a distilled policy are used for model-free actions and planned actions, respectively. Our results on a ecologically relevant, parametric environment indicate that distilled policy network for self-model stabilizes training, has faster inference than using model-free policy, promotes better exploration, and could learn a comprehensive understanding of its own behaviors, at the cost of distilling a new network apart from the model-free policy.

\end{abstract}

\section{Introduction}

% Intro to planning and self model
Planning is an important decision making strategy that allows an agent to search better actions before interacting with the real environment \cite{schrittwieser2020mastering, moravvcik2017deepstack, hafner2019learning}. During planning, it is essential to know which potential actions to evaluate when the action space is large, which is often the case in complex environments \cite{silver2016mastering,silver2017mastering}. Even when the space of atomic actions may be small, the space of high-level actions, required for long-horizon planning, is typically much larger. The actions to explore in planning can be derived by using an approximation of its own decision making strategy: which actions an agent might take in this potential future state. We refer to the model that produces candidate actions during planning as a \textbf{self model} because it approximates potential actions that the agent (the self) may take in real states. If the action space is small, planning could go around by randomly sampling actions from the action space without a self-model. If the action space is large, self-models are necessary to limit the search space of actions \cite{hubert2021learning}.

% Status quo of self model in RL (Works on using model-free policy & dual policy for planning)
While self models are widely used, mostly implicitly, together with world models \cite{ha2018world, hafner2023mastering} to plan actions, it is unclear how self models should be designed. The model-free RL policy, when available, is often used as the self model as it offers reasonable candidate actions to explore in possible future states \cite{silver2016mastering,silver2017mastering,fickinger2021scalable}. While using the model-free policy as the self model is one of the simplest methods, we turn to neuroscience for insights on different self-model designs. Neuroscience studies on reinforcement learning and goal-directed behaviors suggest that the dorsolateral striatum (a part of basal ganglia) is involved in making model-free, habitual actions \cite{packard2002learning, pasupathy2005different, yin2006role} whereas planned actions for goal-directed behaviors activate the prefrontal cortex \cite{daw2005uncertainty, gremel2013orbitofrontal, drummond2020model, cunningham2021dorsolateral}. This raises the question on the implications of having a separate network other than the model-free policy that is specialized for higher level cognitive functions such as planning \cite{russin2020deep}. One previous study designed two separate networks for planned and model-free decision making but the general implications of having dual policies for model-free actions and planning remains to be explored \cite{NIPS20179e82757e}.

% Our work
In this work, we investigate the possibility of having a self-model policy network that is only used for planning, apart from the model-free policy that learns to select actions that maximize future rewards. We explore the benefits and limitations of two distinct self model designs: Having the self model be (1) the model-free policy and (2) a separate distilled policy network. We refer to the agent that uses the model-free policy for planning as the shared policy agent, and the agent that uses the distilled policy for planning as the dual policy agent. Through an ecologically relevant, predator/prey environment that could elucidate the benefit of planning, we show that dual policy agent stabilizes training, has faster inference, explores the environment more, and learns a more comprehensive understanding of its own behaviors.

\section{Hypothesized pros and cons of different self-model implementations}

\begin{figure}
    \begin{subfigure}[b]{0.3\textwidth}
        \centering
        \includegraphics[width=\textwidth]{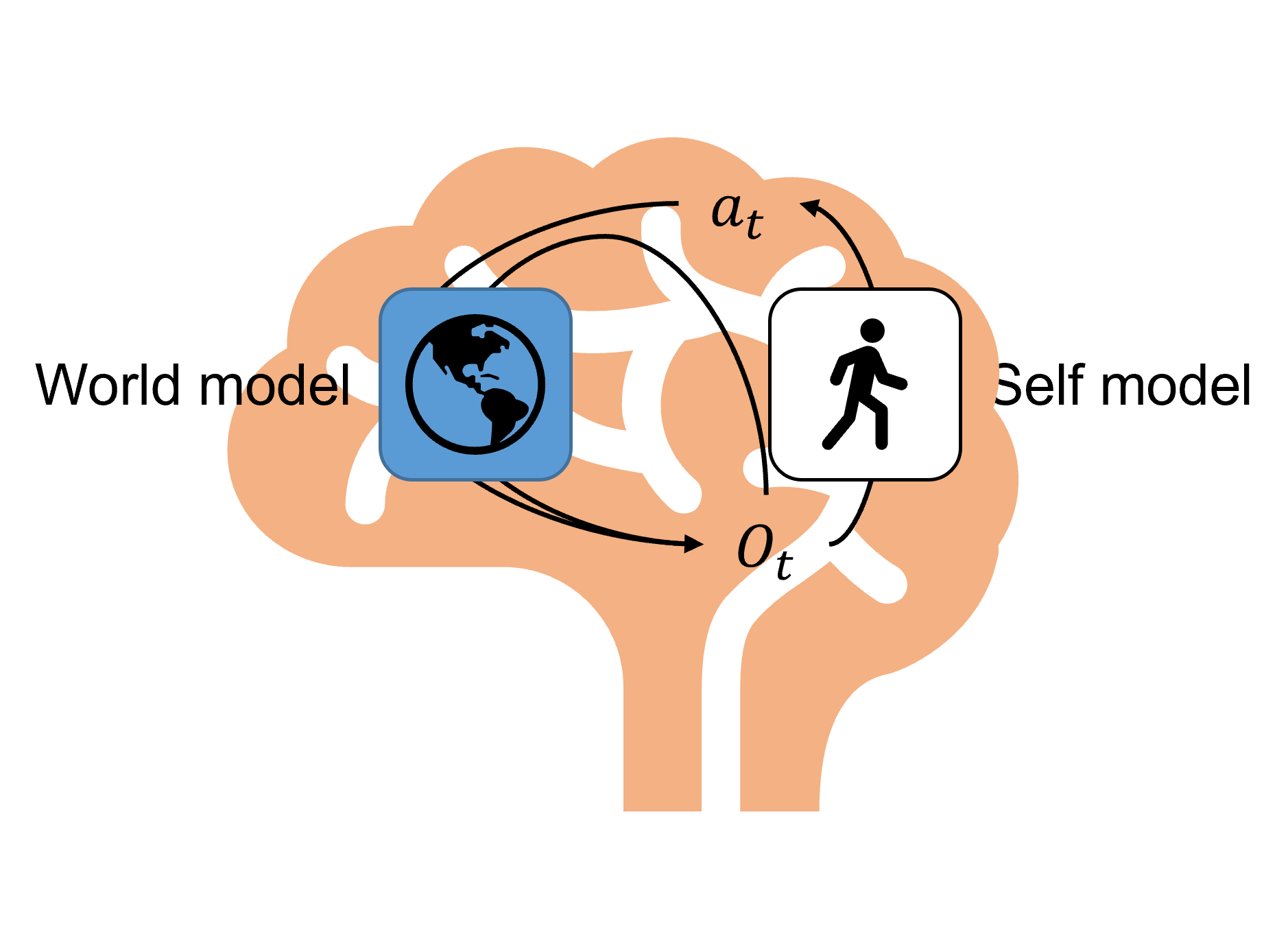}
        \caption{Self model in planning}
        \label{fig:selfmodel}
    \end{subfigure}
    \begin{subfigure}[b]{0.3\textwidth}
        \centering
        \includegraphics[width=\textwidth]{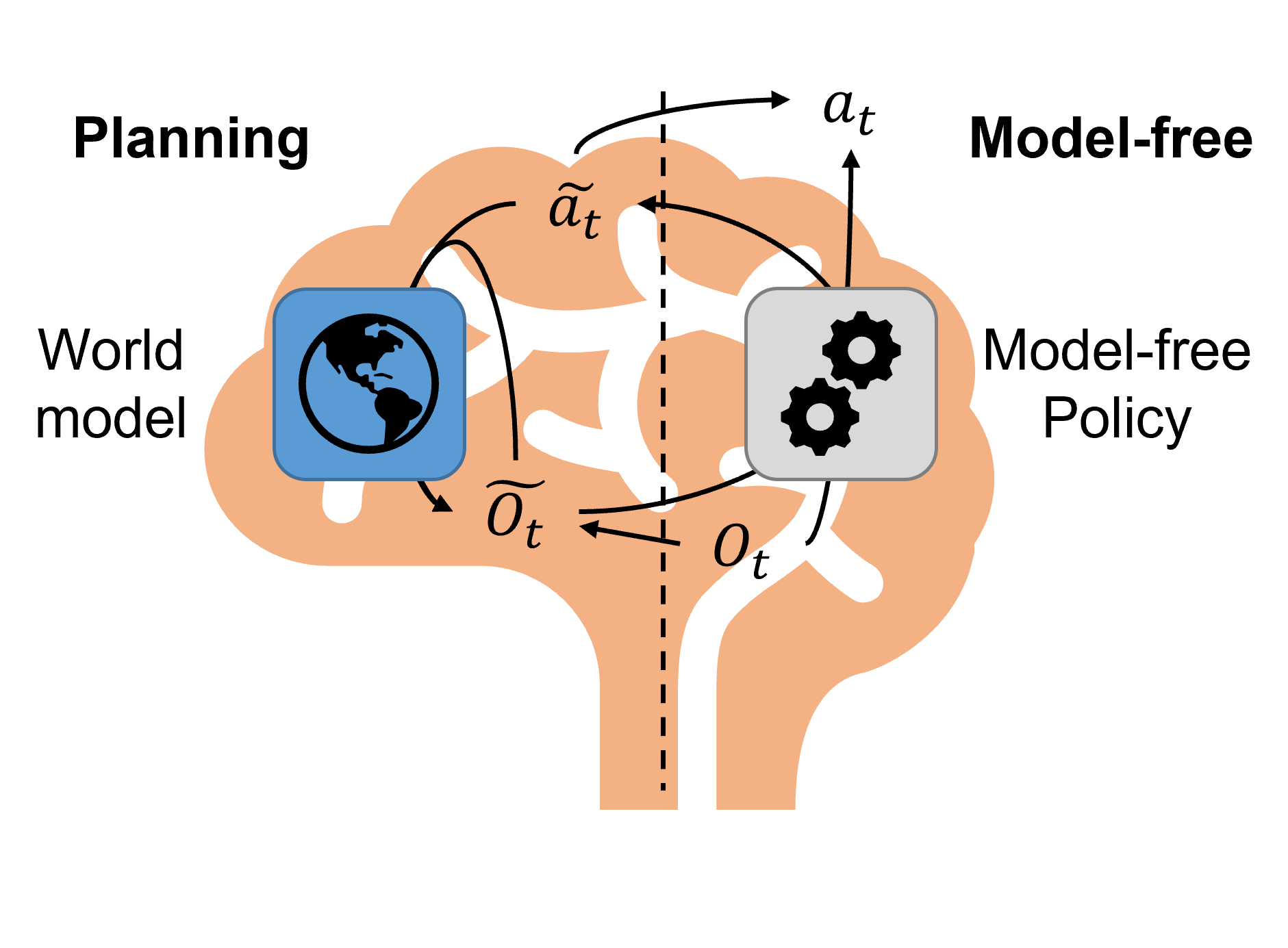}
        \caption{Shared policy agent}
        \label{fig:shared policy agent}
    \end{subfigure}
    \begin{subfigure}[b]{0.3\textwidth}
        \centering
        \includegraphics[width=\textwidth]{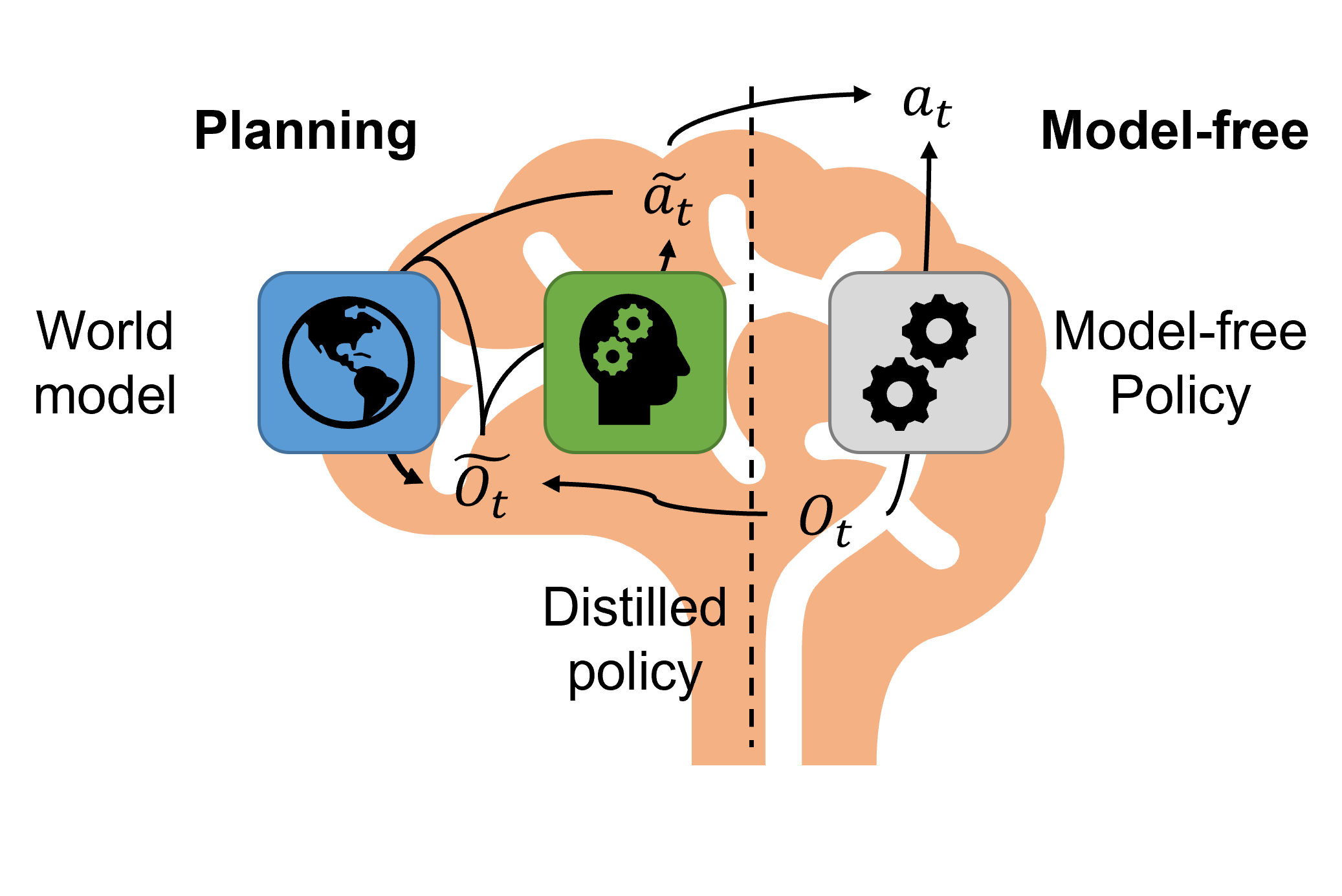}
        \caption{Dual policy agent}
        \label{fig:dual policy agent}
    \end{subfigure}
    \centering
    \caption{Concept of self model and its two distinct designs}
    \label{fig:diagrams}
\end{figure}

\begin{table}
  \caption{Potential benefits of different self model designs}
  \label{sample-table}
  \centering
  \begin{tabular}{ll}
    \toprule
    Shared policy agent & Dual policy agent \\
    \midrule
    Rewarding trajectory & Stable training \\
    Cheaper training & Faster inference for planning \\
     & Better exploration \\
     & Accurate planning \\
    \bottomrule
  \end{tabular}
\end{table}

% Explain our perspective
The concept of self-modeling is not entirely new, with previous studies having explored related, yet distinct ideas. In robotics, an agent's observation encompasses both its internal body states and environmental conditions. To effectively plan interactions with its environment, the agent must predict potential changes to these internal states. Consequently, the agent's predictive ability regarding its body states forms a specialized component of the world model: a model dedicated to predicting observations. This model, which predicts the body states of the agent, has been referred to as the "self model" in recent studies \cite{chen2022fully, sobal2022separating}.

In our study, we explore a distinct concept of the 'self' apart from prediction of one's body states. We define the self-model as a model that generates potential actions for exploration during planning (Figure \ref{fig:selfmodel}). We term it the 'self-model' because it produces actions given an observation, essentially modeling the understanding of one's behavior when simulating future states. Since the self-model receives observations and provides actions, it has been common practice to use the model-free policy as the self-model to sample actions during planning \cite{lee2018deep, gao2017move, schrittwieser2020mastering}. Building upon the neurobiological insight that separate brain regions are responsible for habitual and goal-directed actions \cite{daw2005uncertainty, NIPS20179e82757e}, we explore the implications of having two distinct policy networks for model-free action and planning.

In this dual-network setup, one network learns to predict the agent's actions by observing its past actions, similar to how a world model learns to forecast the future. We refer to this network as the distilled policy network. In contrast, the other network, the model-free policy, aims to predict actions that would most likely lead to rewarding trajectories. This dual policy agent uses its model-free policy for model-free actions, and its distilled policy as the self model for planning (Figure \ref{fig:dual policy agent}). This is in contrast to the shared policy agent which contains only the model-free policy and uses it for both action types (Figure \ref{fig:shared policy agent}).

We hypothesize advantages and limitations for both the shared policy and dual policy agents. For the shared policy agent, the key benefit is that model-free policy could identify rewarding trajectories that lead to more rewarding planned actions, which could enhance overall performance. Additionally, there's no necessity for training an extra network, preventing the introduction of additional computational burden. On the other hand, the dual policy agent potentially offers a set of unique advantages. Firstly, having multiple networks may stabilize training for planning, as previous works with multiple networks such as ensemble models \cite{fort2019deep, osband2016deep} and double Q networks \cite{van2016deep, fujimoto2018addressing} showed improved stability for model-free learning. Secondly, the inference time for planning can be faster if distilled policy is smaller than the model-free policy, as the dual policy allows flexible network design. Another advantage is the potential for better exploration suggested by a distilled policy with higher entropy \cite{NIPS20179e82757e}. Lastly, the agent's decision-making system might consist of circuits other than the model-free policy, which override in different conditions such as reflexive behaviors \cite{flacco2012depth, yasin2020unmanned, fanselow1994neural, ledoux2000emotion}. To plan accurately, a holistic understanding of one's decision-making strategy is required, which cannot be achieved solely by the model-free policy. This necessitates a separate model to oversee past behaviors across different states, further supporting the use of a dual policy agent.

\section{Experiment setup}

\subsection{Environment}

\begin{figure}
    \begin{subfigure}[b]{0.3\textwidth}
        \centering
        \includegraphics[width=\textwidth]{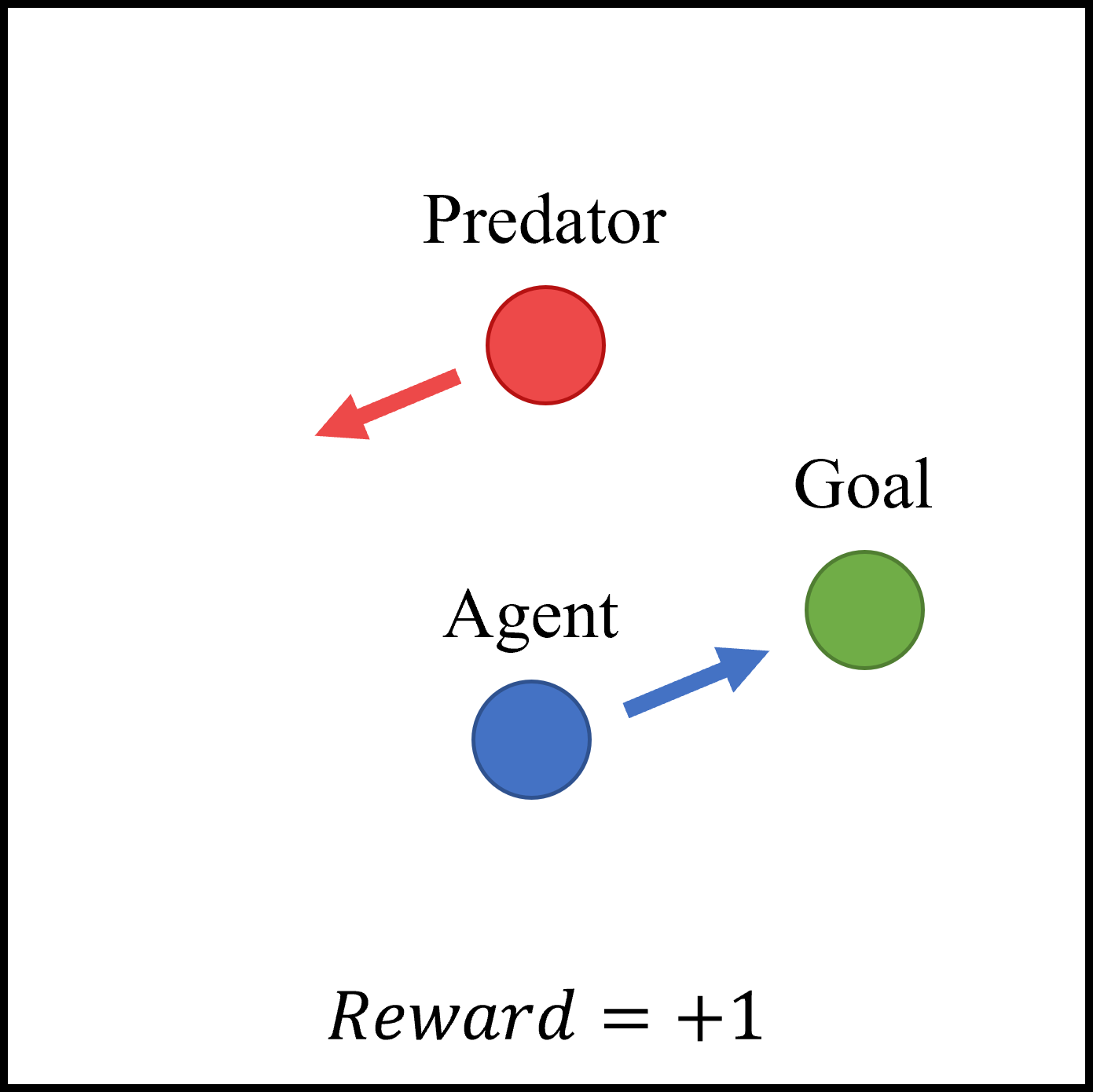}
        \caption{Success}
    \end{subfigure}
    \begin{subfigure}[b]{0.3\textwidth}
        \centering
        \includegraphics[width=\textwidth]{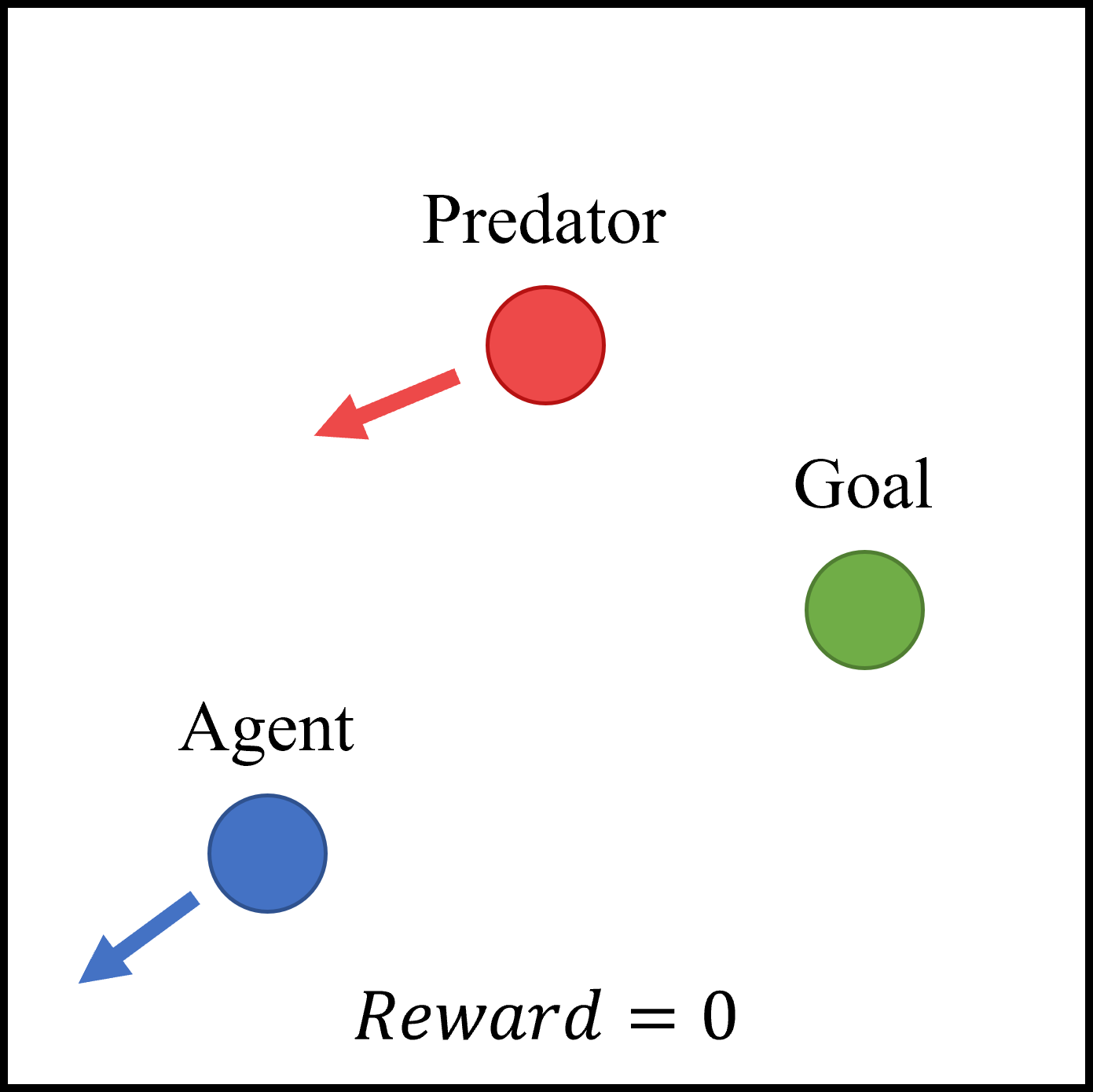}
        \caption{Timeout}
    \end{subfigure}
    \begin{subfigure}[b]{0.3\textwidth}
        \centering
        \includegraphics[width=\textwidth]{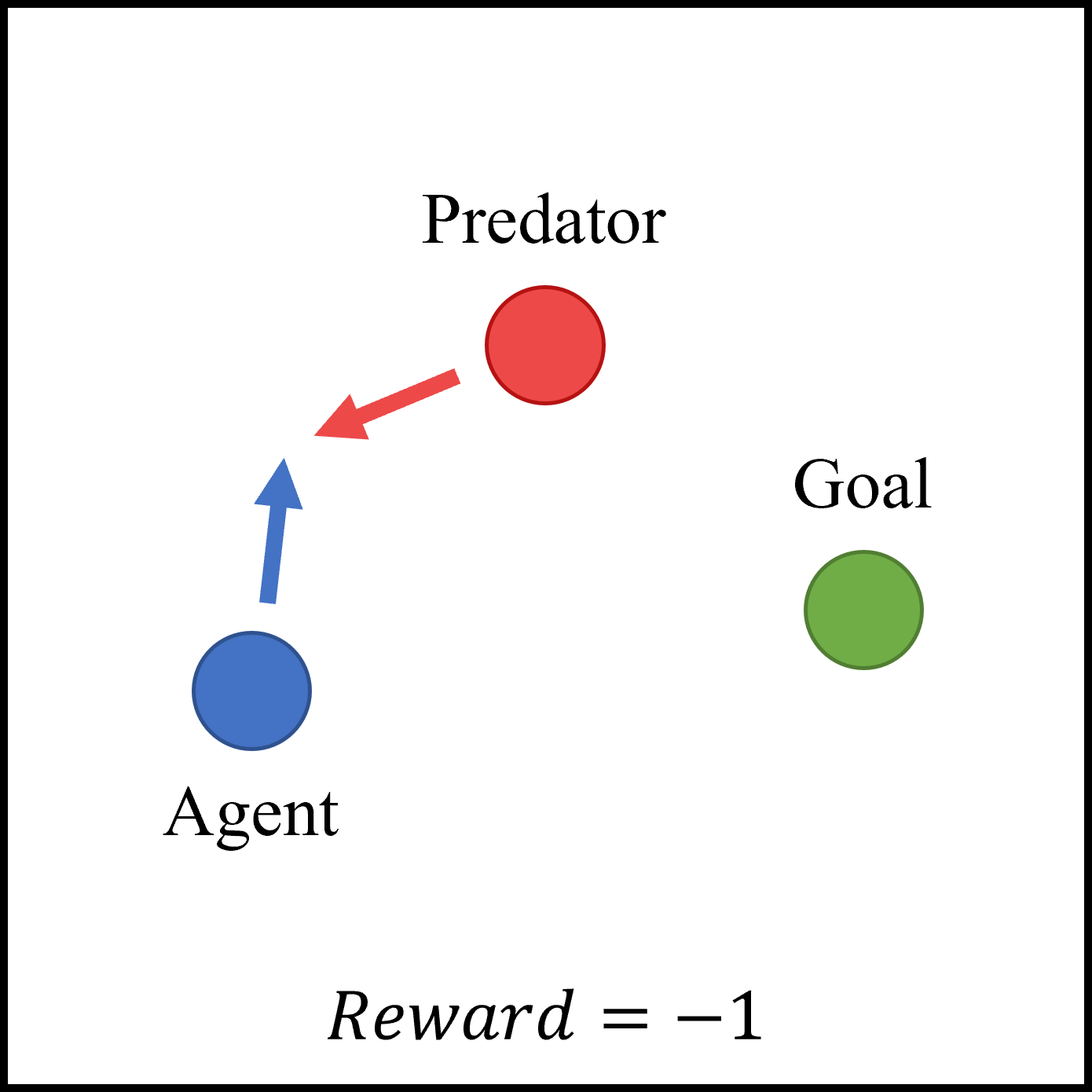}
        \caption{Death}
    \end{subfigure}
    \centering
    
    \caption{Environment reward settings}
    \label{fig:my_label}
\end{figure}

% What do we desire in this environment (particle world style)
% High-dimensional action space
% Planning to be useful
% Parametric control of the environment
% Check Particle world for environment justificaiton
We evaluate our hypotheses in a survival setting where the agent must evade the predator while navigating toward the goal in a bounded box environment, similar to the environment proposed in \cite{lowe2017multi}. This survival environment serves as a testbed to assess the advantages of planning over trial-and-error approaches, as the consequences of negative rewards can be catastrophic, such as death \cite{mugan2020spatial}. The action space of the environment is continuous which necessitates the self model as the search space for actions during planning is large. The environment is parameterized and we test our agents in varying environment settings. The details of the environment design is described in Supplementary Materials.

% Planning
For planning in shared policy agent and dual policy agent, Monte Carlo Tree Search (MCTS) \cite{kocsis2006bandit,coulom2007efficient,browne2012survey} is performed. Future states are simulated using the world model $o_{t+1}, r_{t+1} = f(o_t, a_t)$ and the self model $a_t, v_t = g(o_t)$ starting from the current observation, where $o_t$, $r_t$, $a_t$, and $v_t$ refers to observation, reward, action, and value of the observation, respectively. At the initial observation for planning, four action candidates are randomly sampled with one action being the mean action from the action distribution generated from the self model. After the initial action candidates, mean action from the action distribution are always taken to explore possible future states, which results in a sparse tree search. This was to simulate most rewarding trajectories after randomly selecting initial actions for exploration. The MCTS was performed until maximum depth of four. To evaluate different trajectories, advantage of each trajectory was measured using generalized advantage estimation (GAE) \cite{schulman2015high} from the reward of future trajectories created by the world model. For details on planning and its hyperparameters, see Supplementary Materials.

\subsection{Networks}
% Network architecture
In our setup, we utilize three neural networks: a model-free policy network $a_t=\pi_{\theta}(o_t)$ and $v_t=V^{\pi}(o_t)$, a world model network $o_{t+1}, r_{t+1} = f_{\theta}(o_t, a_t)$, and a distilled policy network $a_t=\tilde{\pi}_{\theta}(o_t)$ and $v_t=V^{\tilde{\pi}}(o_t)$. The model-free policy network $\pi_{\theta}$ takes the current environmental observation $o_t$ as input, which includes the spatial coordinates of the agent, the goal, and the predator. From this, it predicts the future direction $dx, dy$ the agent should move towards. It also contains a separate value prediction network $V^{\pi}$ that predicts the value $v_t$ of the given observation.

The world model network $f_{\theta}$ receives a history of past observations and the agent's actions, then predicts the next state of the environment and the associated reward. The input to this network includes the coordinates of the agent, goal, and predator from the current and previous timestep, as well as the distances between these entities. We have incorporated observations of two timesteps and distance values for the world model to facilitate learning, as our primary focus lies in the design of self models.

The distilled policy network $\tilde{\pi}_{\theta}$ receives the current environmental observation $o_t$ and predicts the action $a_t$ that the agent previously took. Additionally, it predicts the value $v_t$ of the current observation. In contrast to the model-free policy network, which uses separate networks for action and value prediction, the distilled policy network combines these predictions into a single network. This design was selected to examine performance of the distilled policy under harder conditions, as having separate specialized networks for action and value prediction would perform better. This design also makes the distilled policy smaller than the model-free policy network for the same hidden layer configuration, which may benefit inference time for planning. The details of network architecture is described in the Supplementary Materials.

% Learning algorithm
The model-free policy is trained using proximal policy optimization (PPO) algorithm \cite{schulman2017proximal} using experience rollouts every predefined interval (Minimize $\mathcal{L}_{PPO}$, Eq. \ref{eq: PPO loss}). The advantage $\hat{A}_t$ was computed using the GAE. The world model is trained to predict next step observation and reward by receiving current observation and action. MSE Loss $\mathcal{L}_{World model}$ between the $o_{t+1}$, $r_{t+1}$ and $\tilde{o}_{t+1}$, $\tilde{r}_{t+1}$ is used to optimize the network, where $o_{t+1}$, $r_{t+1}$ refers to real experience rollouts, and $\tilde{o}_{t+1}$, $\tilde{r}_{t+1}$ refers to world model predictions (Eq. \ref{eq: World model loss}). The distilled policy is trained to predict past history of actions given an observation. To maximize the probability of past actions given the action probability distribution predicted from the distilled policy, the distilled policy minimizes the negative log-likelihood $\mathcal{L}_{Action}$ (Eq. \ref{eq: Negative log likelihood}). The distilled policy also predicts the value $v_t$ of the given state, and is optimized to minimize the knowledge distillation loss $\mathcal{L}_{KD}$ \cite{hinton2015distilling, van2020brain} given past observation and value (Eq. \ref{eq: Knowledge distillation loss}). The $T$ refers to the temperature parameter of the soft cross entropy loss. The hyperparameters for training are described in Supplementary Materials.

\begin{equation}
\mathcal{L}_{PPO} = \mathbb{E} \left[ \min \left( \frac{p(a_t|\pi_{\theta}(o_t))}{p(a_t|\pi_{\theta_{old}}(o_t))} \hat{A}_t, clip\left( \frac{p(a_t|\pi_{\theta}(o_t))}{p(a_t|\pi_{\theta_{old}}(o_t))} , 1 - \epsilon, 1 + \epsilon\right)\hat{A}_t\right) \right]
\label{eq: PPO loss}
\end{equation}

\begin{equation}
\mathcal{L}_{World model} = \mathbb{E} \left[ \left( \tilde{o}_{t+1} - o_{t+1} \right)^2 + \left( \tilde{r}_{t+1} - r_{t+1} \right)^2 \right]
\label{eq: World model loss}
\end{equation}

\begin{equation}
\mathcal{L}_{Action} = \mathbb{E} \left[ - \log p(a_t| \tilde{\pi_{\theta}}(o_t)) \right]
\label{eq: Negative log likelihood}
\end{equation}

\begin{equation}
\mathcal{L}_{KD} = -\mathbb{E}\left[ T^2 \sum_{i=1}^{K} V^{\pi}(o_t) \log V^{\tilde{\pi}}(o_t) \right]
\label{eq: Knowledge distillation loss}
\end{equation}

\section{Results}

\subsection{Dual policy agent achieves similar performance to shared policy agent with better stability}
\label{subsection: stability}

\begin{table}[h]
  \caption{Neural networks of agents}
  \label{tab: agent modules}
  \centering
  \begin{tabular}{lccc}
    \toprule
    & Model-free policy & World model & Distilled policy \\
    \midrule
    Simple agent & O & X & X \\
    Shared policy agent & O & O & X \\
    Dual policy agent & O & O & O \\
    \bottomrule
  \end{tabular}
\end{table}

\begin{figure}[h]
    \centering
    \begin{subfigure}[b]{0.3\textwidth}
        \centering
        \includegraphics[width=\textwidth]{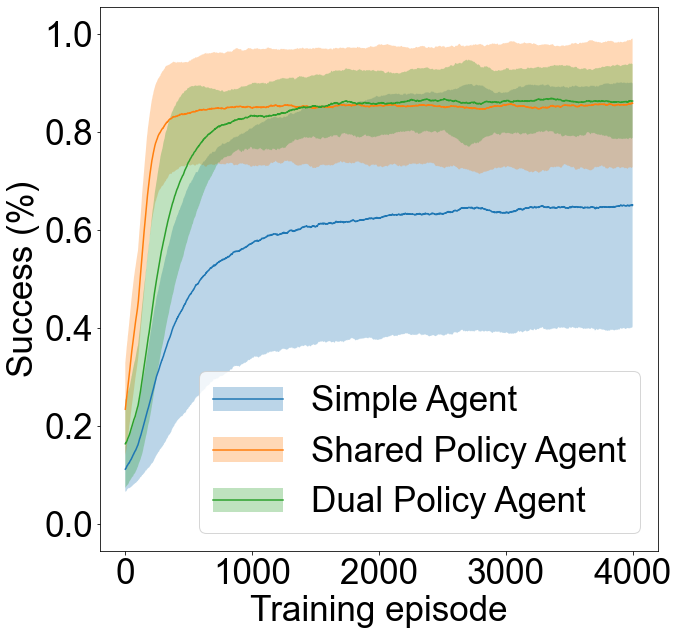}
        \caption{}
    \end{subfigure}
    \begin{subfigure}[b]{0.3\textwidth}
        \centering
        \includegraphics[width=\textwidth]{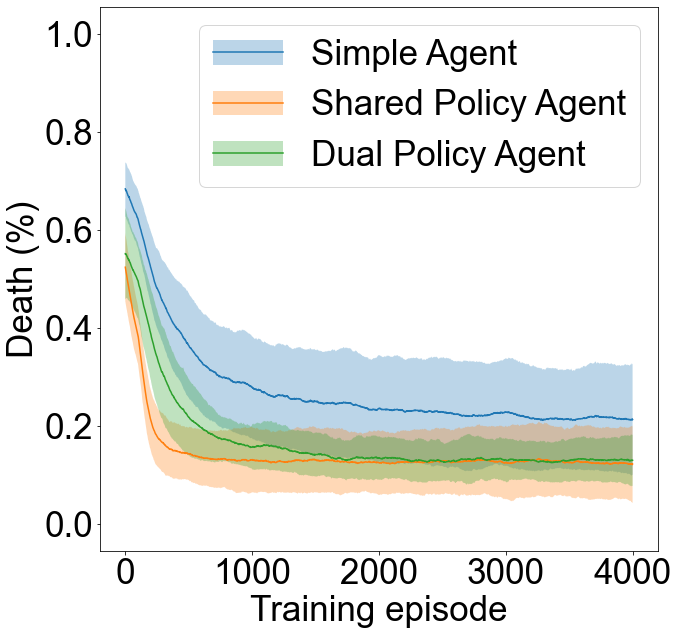}
        \caption{}
    \end{subfigure}
    \begin{subfigure}[b]{0.3\textwidth}
        \centering
        \includegraphics[width=\textwidth]{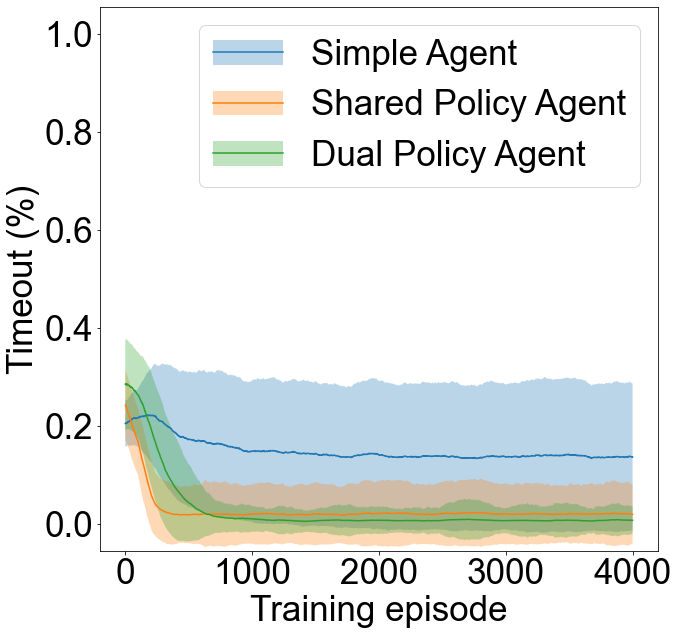}
        \caption{}
    \end{subfigure}
    \caption{Proportions of episode outcomes during training episodes. Simple, shared policy, and dual policy agent are compared. The shaded area is standard deviation.}
    \label{fig: training}
\end{figure}

\begin{table}[h]
  \caption{Performance of agents during evaluation episodes (Mean$\pm$std\%).}
  \label{tab: evaluation}
  \centering
  \begin{tabular}{lccc}
    \toprule
    & Proportion of success & Proportion of death & Proportion of timeout \\
    \midrule
    Simple agent & 65.76$\pm$25.35 & 20.54$\pm$10.82 & 13.70$\pm$15.29 \\
    Shared policy agent & 85.61$\pm$13.21 & 12.26$\pm$7.34 & 2.12$\pm$6.54 \\
    Dual policy agent & 86.87$\pm$\textbf{6.93} & 12.50$\pm$\textbf{4.35} & 0.62$\pm$\textbf{2.89} \\
    \bottomrule
  \end{tabular}
\end{table}

% Agent, model
Three different agents are compared to asses the characteristics of different self model designs (Table \ref{tab: agent modules}). The simple agent only contains a model-free policy and only employs model-free actions. The simple agent acts as a lower bound to demonstrate the performance benefit of planning. The shared policy agent contains a model-free policy and the world model. The agent and employs both model-free action and planned actions where it uses its model-free policy as the self-model to sample candidate actions during planning. The switch between the two decision making strategies is determined randomly. The dual policy agent contains the world model, model-free policy, and the distilled policy. The agent uses its model-free policy for model-free actions, and its distilled policy as the self-model for sampling actions for planning.

We measured the proportion of each outcome types to evaluate the performance of our agents. Figure \ref{fig: training} reveals that the dual policy agent exhibits similar performance to the shared policy agent, suggesting that the distilled policy sufficiently learns actions that lead to rewarding trajectories. While the mean performance for the dual policy agent and the shared policy agent is similar for all three reward types, the dual policy agent shows smaller standard deviation which indicates more stable training compared to the shared policy agent. This could also be found from evaluation results, where the standard deviation of the dual policy agent were the smallest for all three reward types (Table \ref{tab: evaluation}). While the dual policy agent outperforms the shared policy agent, the difference wasn't statistically significant from our experiments (p=0.403, 0.778, 0.039 for success, death, and timeout, respectively. Two sample t-test). The simple agent showed the minimum performance among the agents which highlights the benefit of planning. We further report experiment results of our dual policy agent with different self model designs in the Supplementary Materials.

\subsection{Dual policy agent have faster planning inference with smaller distilled networks}
\label{subsection: cost}

\begin{table}[h]
  \caption{Number of parameters in each policy network}
  \label{tab: number of parameters}
  \centering
  \begin{tabular}{lll}
    \toprule
    Number of hidden layer neurons & Model-free policy & Distilled policy \\
    \midrule
    $\left[32\right]$ & 549 & 325 \\
    $\left[64,64\right]$ & 9,413 & 4,805 \\
    $\left[128,128,128,128\right]$ & 101,253 & 50,821 \\
    \bottomrule
  \end{tabular}
\end{table}

\begin{figure}[h]
  \centering
  \begin{subfigure}[b]{0.3\textwidth}
         \centering
         \includegraphics[width=\textwidth]{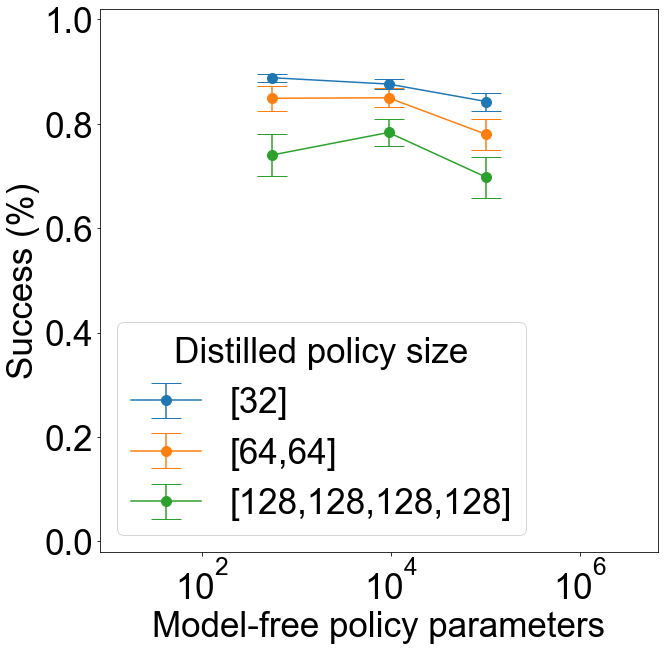}
         \caption{}
     \end{subfigure}
    \begin{subfigure}[b]{0.3\textwidth}
         \centering
         \includegraphics[width=\textwidth]{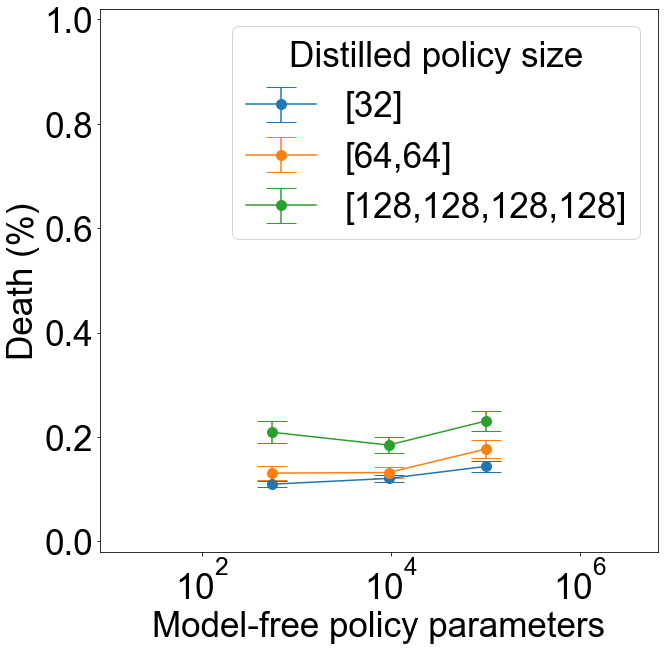}
         \caption{}
     \end{subfigure}
     \begin{subfigure}[b]{0.3\textwidth}
         \centering
         \includegraphics[width=\textwidth]{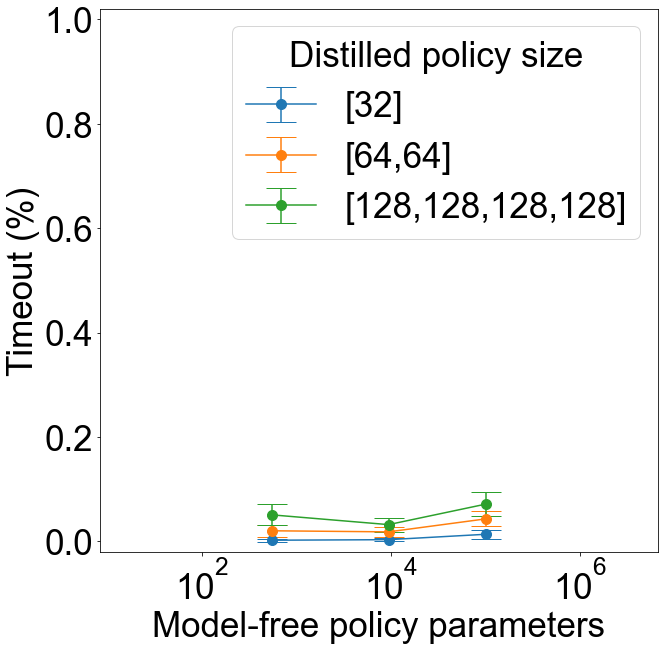}
         \caption{}
     \end{subfigure}
  \caption{Proportions of rewards during evaluation episodes across different model-free and distilled policy sizes. Error bars indicate 95\% confidence intervals.}
  % \caption{Reward type distribution across different model-free and distilled policy network architectures. Letters S, M, and L stand for small, medium, and large networks, respectively.}
  \label{fig: cost}
\end{figure}

In the process of planning, actions are sampled from the self model for each future state that is being explored. Therefore, a self model with faster inference capability can accelerate the planning process. Given this, we sought to explore the relationship between various architectures of the model-free policy and the distilled policy within the framework of the dual policy agent as the number of parameters determines the amount of computation required for inference. We explored three difference architecture settings for the model-free policy and the distilled policy (Table \ref{tab: number of parameters}).

Our results showed a consistent performance enhancement with smaller distilled policies, irrespective of the size of the model-free policy (Figure \ref{fig: cost}). This suggests that smaller networks are more efficient at distilling the knowledge from the model-free policy. This finding is advantageous as it implies that the distilled policy can be designed to be smaller than the model-free policy, thereby offering not only a faster inference but also superior performance. In this environment, we did not observe any discernible pattern correlating the size of the model-free policy with performance.

\subsection{Dual policy enhances exploration}

\begin{figure}[h]
  \centering
    \begin{subfigure}[b]{0.3\textwidth}
         \centering
         \includegraphics[width=\textwidth]{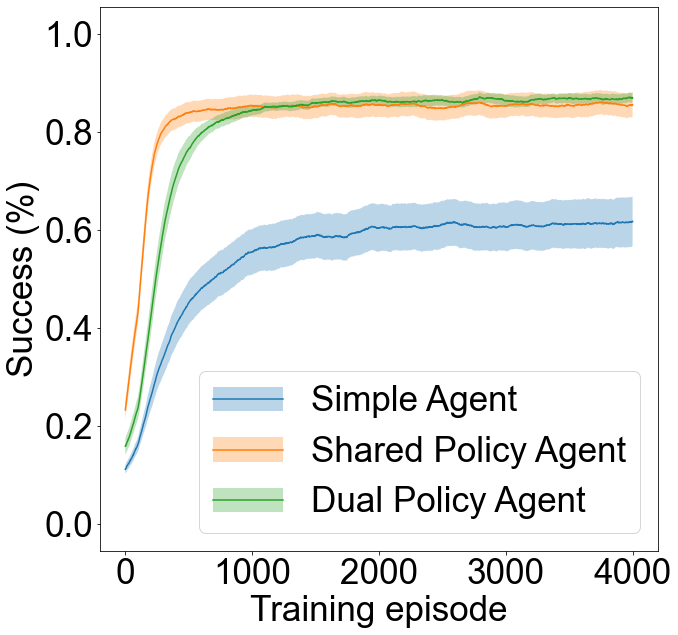}
         \caption{}
     \end{subfigure}
    \begin{subfigure}[b]{0.3\textwidth}
         \centering
         \includegraphics[width=\textwidth]{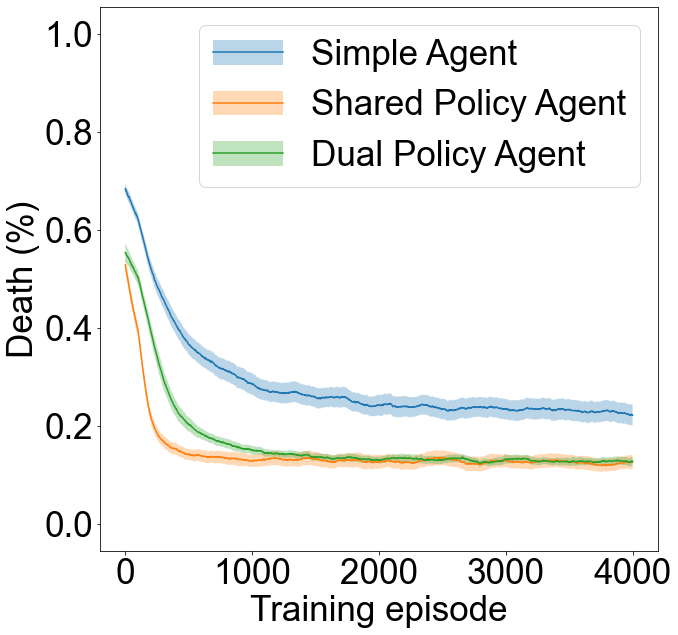}
         \caption{}
     \end{subfigure}
     \begin{subfigure}[b]{0.3\textwidth}
         \centering
         \includegraphics[width=\textwidth]{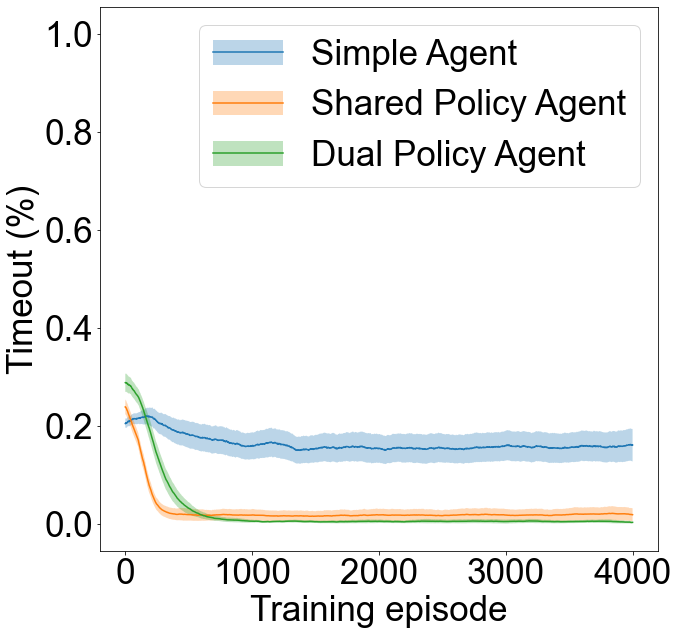}
         \caption{}
     \end{subfigure}
     \begin{subfigure}[b]{0.3\textwidth}
         \centering
         \includegraphics[width=\textwidth]{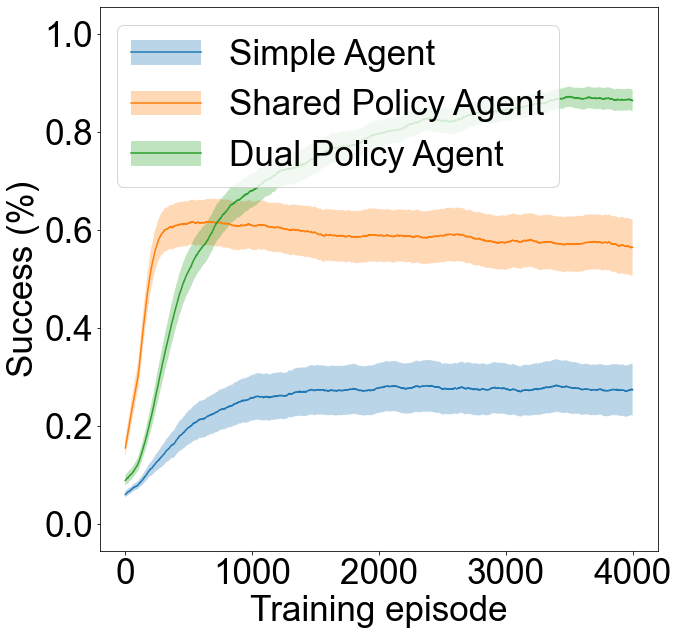}
         \caption{}
     \end{subfigure}
     \begin{subfigure}[b]{0.3\textwidth}
         \centering
         \includegraphics[width=\textwidth]{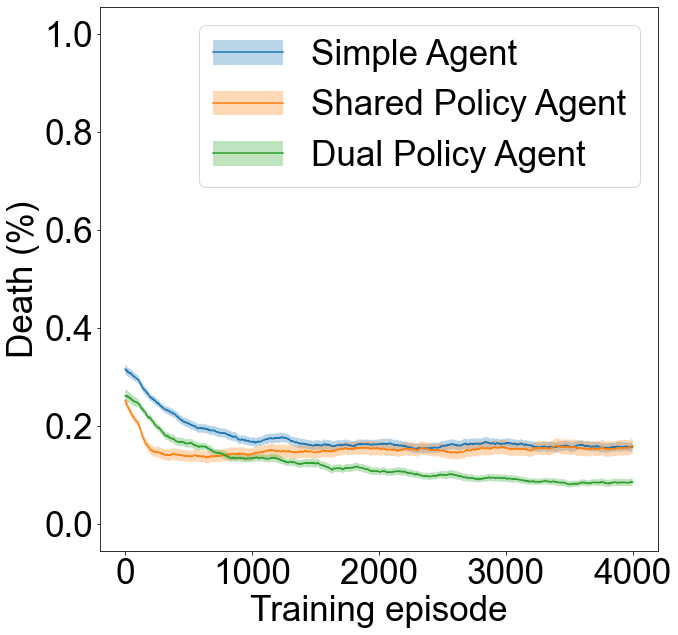}
         \caption{}
     \end{subfigure}
     \begin{subfigure}[b]{0.3\textwidth}
         \centering
         \includegraphics[width=\textwidth]{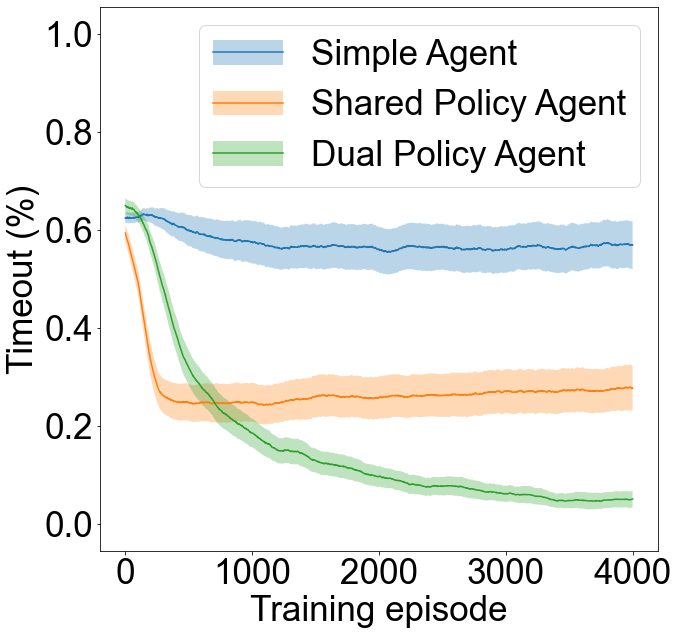}
         \caption{}
     \end{subfigure}
     \begin{subfigure}[b]{0.3\textwidth}
         \centering
         \includegraphics[width=\textwidth]{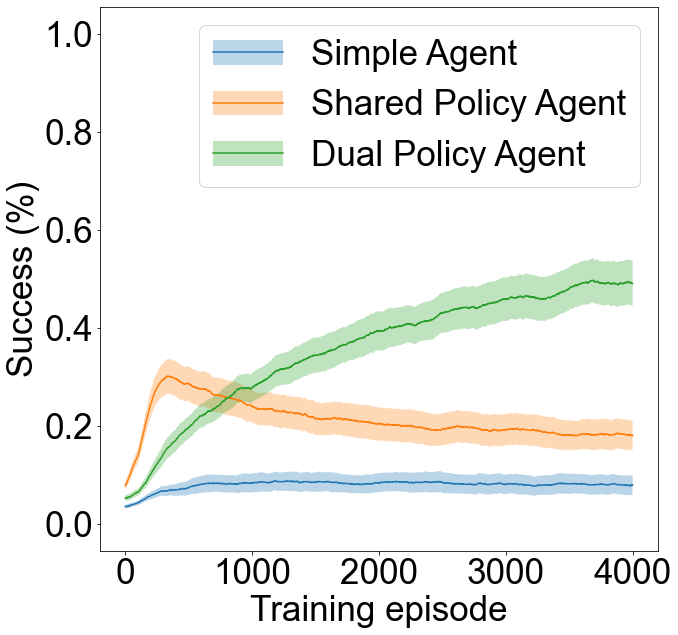}
         \caption{}
     \end{subfigure}
     \begin{subfigure}[b]{0.3\textwidth}
         \centering
         \includegraphics[width=\textwidth]{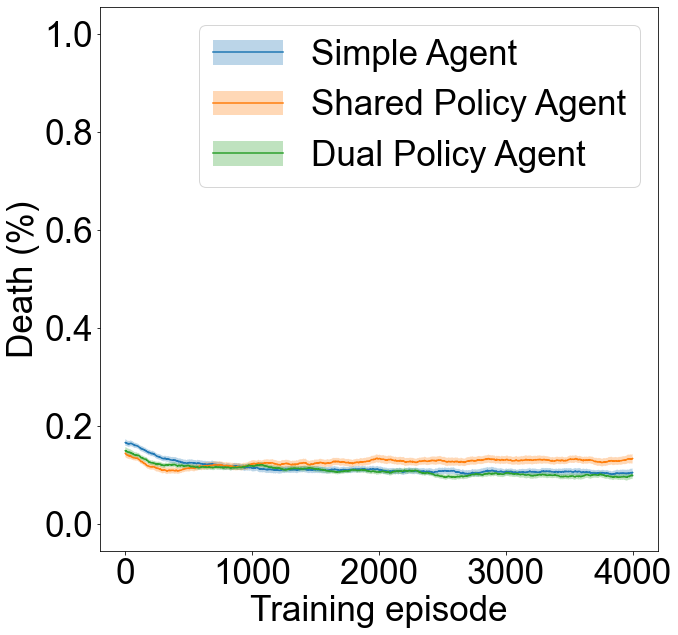}
         \caption{}
     \end{subfigure}
     \begin{subfigure}[b]{0.3\textwidth}
         \centering
         \includegraphics[width=\textwidth]{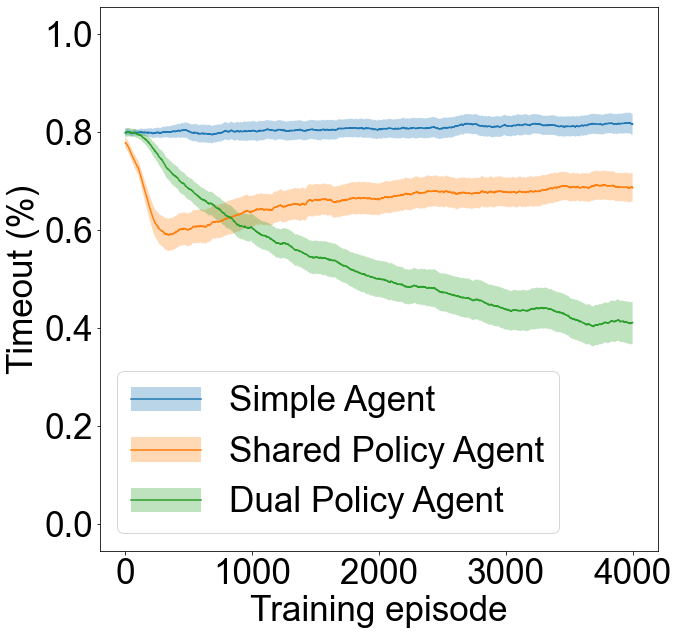}
         \caption{}
     \end{subfigure}     
  \caption{Proportions of reward types during training episodes for different map sizes. (a)-(c): Map size of 10 units, (d)-(f): Map size of 20 units, (g)-(i): Map size of 30 units. Simple, shared policy, and dual policy agent are compared. The shaded area is 95\% confidence interval.}
  \label{fig: training_exploration}
\end{figure}

\begin{table}[h]
  \caption{Success proportion of agents during evaluation episodes. Values indicate mean and 95\% confidence interval.}
  \label{tab: evaluation_exploration}
  \centering
  \begin{tabular}{lccc}
    \toprule
    & \multicolumn{3}{c}{Map size} \\
    & 10 units & 20 units & 30 units \\
    \midrule
    Simple agent & 62.46 (57.30, 67.62) & 27.52 (22.26, 32.77) & 7.74 (5.78, 9.70) \\
    Shared policy agent & 86.16 (83.73, 88.59) & 56.36 (50.66, 62.06) & 18.33 (15.30, 21.36) \\
    Dual policy agent & 87.55 (86.80, 88.29) & 87.39 (85.35, 89.43) & 49.32 (44.59, 54.05) \\
    \bottomrule
  \end{tabular}
\end{table}

The model-free policy often finds itself confined to a singular rewarding trajectory. Previous research employing dual policy as self-model for planning \cite{NIPS20179e82757e} suggested that one of the distilled policy's advantages is its facilitation of exploration in real environments, thereby accruing a diverse range of experience rollouts. To evaluate the influence of exploration on performance, we subjected our simple, shared policy, and dual policy agents to environments of varying map sizes. Larger maps inherently demand more exploration, providing a more challenging test of the agents' exploration capabilities.

We found consistent results with the previous study, where the dual policy agent outperforms the shared policy agent drastically as the map size increases. The success proportion of dual policy agent outperforms the shared policy agent during training and evaluation episodes (Figure \ref{fig: training_exploration}, Table \ref{tab: evaluation_exploration}). The timeout proportion of the dual policy agent is also significantly smaller than the shared policy agent which indicates the agent is able to navigate the goal in a more precise manner. The simple agent showed the minimum performance among the agents which highlights the benefit of planning again.

\subsection{Dual policy learns comprehensive understanding of the agent's behaviors}

\begin{figure}[h]
  \centering
  \begin{subfigure}[b]{0.3\textwidth}
         \centering
         \includegraphics[width=\textwidth]{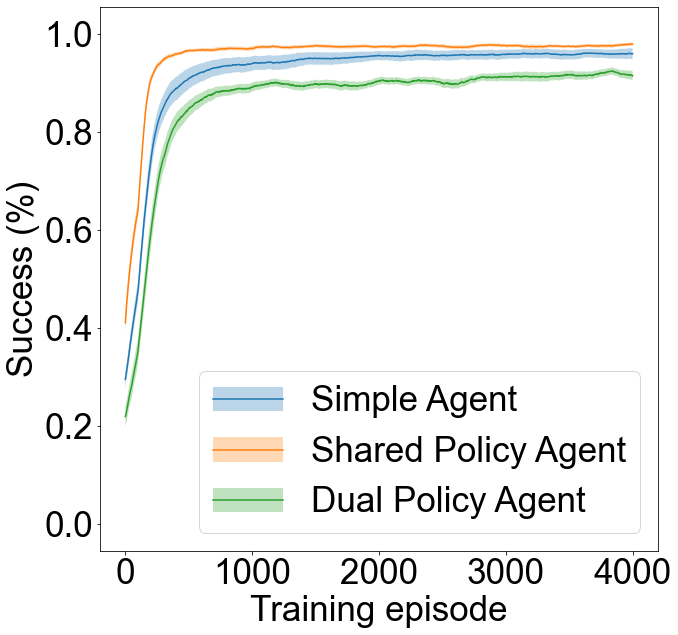}
         \caption{}
     \end{subfigure}
    \begin{subfigure}[b]{0.3\textwidth}
         \centering
         \includegraphics[width=\textwidth]{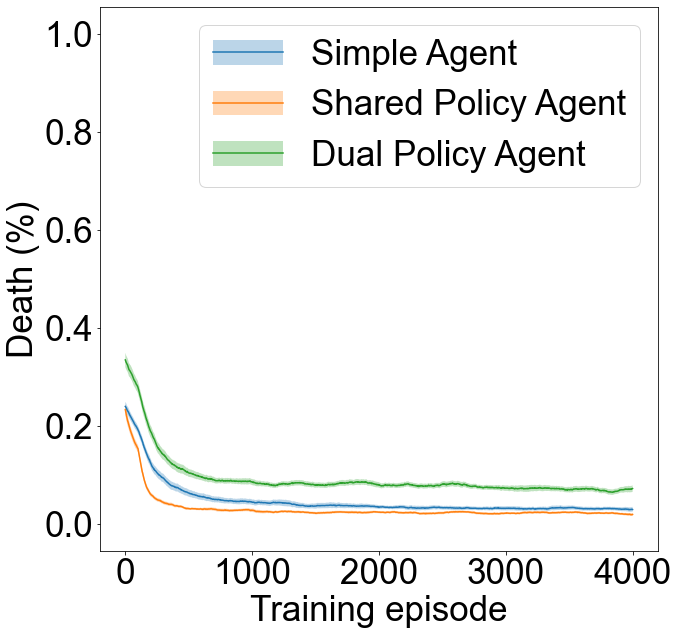}
         \caption{}
     \end{subfigure}
     \begin{subfigure}[b]{0.3\textwidth}
         \centering
         \includegraphics[width=\textwidth]{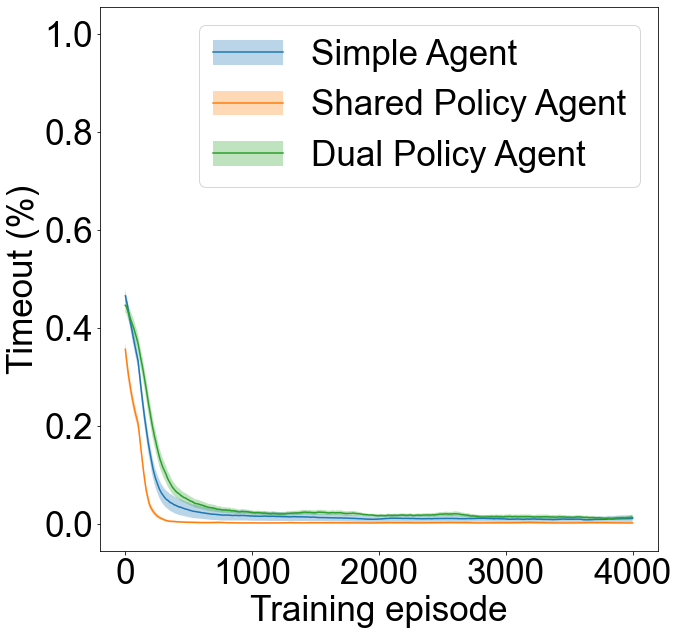}
         \caption{}
     \end{subfigure}
  \begin{subfigure}[b]{0.3\textwidth}
         \centering
         \includegraphics[width=\textwidth]{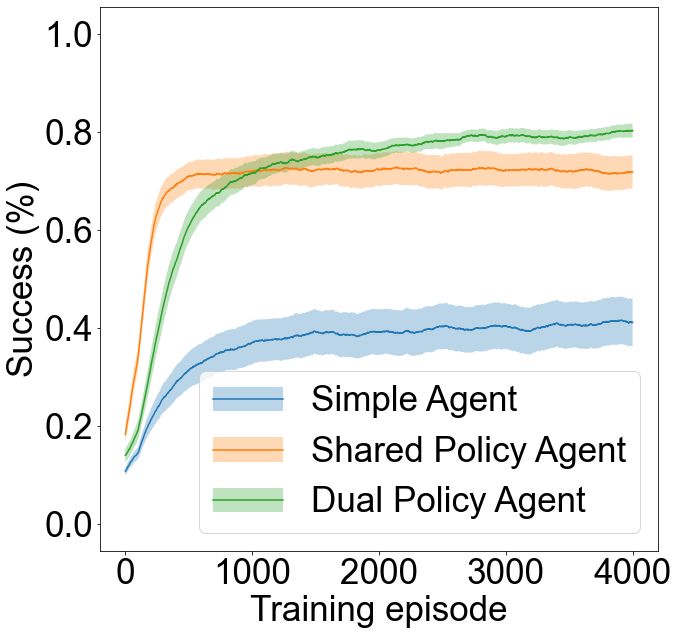}
         \caption{}
     \end{subfigure}
    \begin{subfigure}[b]{0.3\textwidth}
         \centering
         \includegraphics[width=\textwidth]{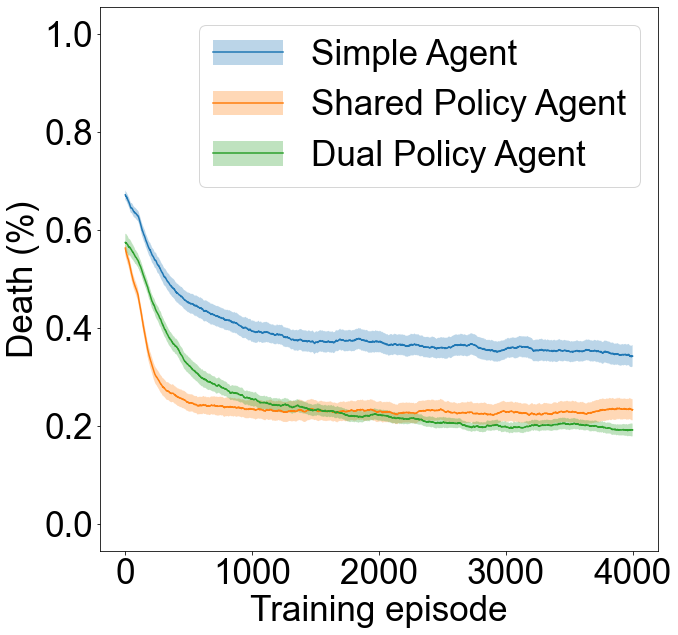}
         \caption{}
     \end{subfigure}
     \begin{subfigure}[b]{0.3\textwidth}
         \centering
         \includegraphics[width=\textwidth]{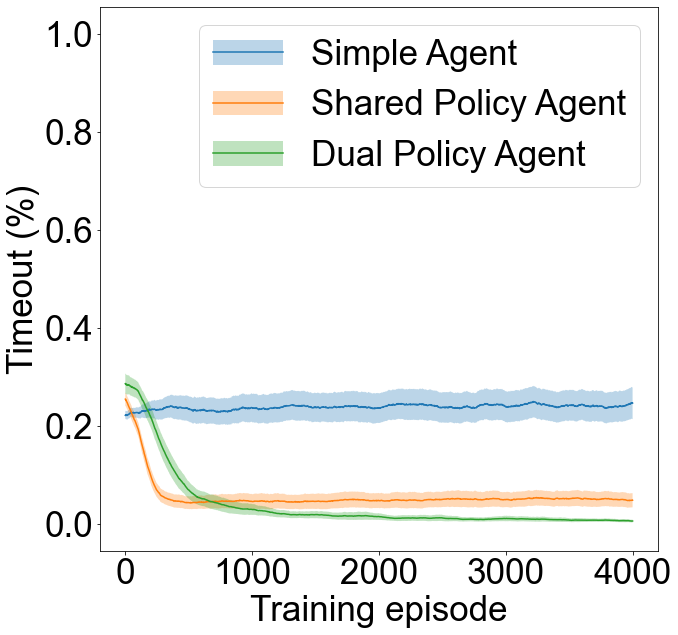}
         \caption{}
     \end{subfigure}
  \caption{Proportions of reward types during training episodes. Simple, shared policy, and dual policy agent with reflexive behavior are compared. (a)-(c): flight reflexive behavior, (d)-(f): freeze reflexive behavior. The shaded area is 95\% confidence interval.}
  \label{fig: training_innate}
\end{figure}

\begin{table}[h]
  \caption{Performance of agents with flight reflexive behavior during evaluation episodes. Values indicate mean and 95\% confidence interval.}
  \label{tab: evaluation_innate_flight}
  \centering
  \begin{tabular}{lccc}
    \toprule
    & Proportion of success & Proportion of death & Proportion of timeout \\
    \midrule
    Simple agent & 97.87 (96.03, 97.71) & 2.36 (2.03, 2.68) & 0.78 (0.23, 1.32) \\
    Shared policy agent & 97.83 (97.44, 98.23) & 1.98 (1.74, 2.21) & 0.19 (0.01, 0.38)\\
    Dual policy agent & 92.60 (91.82, 93.39) & 6.41 (5.92, 6.90) & 0.99 (0.66, 1.32) \\
    \bottomrule
  \end{tabular}
\end{table}

\begin{table}[h]
  \caption{Performance of agents with freeze reflexive behavior during evaluation episodes. Values indicate mean and 95\% confidence interval.}
  \label{tab: evaluation_innate_freeze}
  \centering
  \begin{tabular}{lccc}
    \toprule
    & Proportion of success & Proportion of death & Proportion of timeout \\
    \midrule
    Simple agent & 41.68 (36.75, 46.61) & 33.69 (31.56, 35.83) & 24.62 (21.46, 27.79) \\
    Shared policy agent & 72.35 (68.92, 75.79) & 22.77 (20.59, 24.96) & 4.87 (3.34, 6.40) \\
    Dual policy agent & \textbf{80.73 (79.46, 82.00)} & \textbf{18.61 (17.64, 19.58)} & \textbf{0.66 (0.27, 1.05)} \\
    \bottomrule
  \end{tabular}
\end{table}

Agents might possess parts of action policies that they cannot simulate for planning purposes, such as reflexive and other innate responses to emergency scenarios \cite{flacco2012depth, yasin2020unmanned}. In biological organisms, these reflexive responses are often genetically hard-wired, and rely on separate circuitry from those for habitual, learned policy \cite{fanselow1994neural, ledoux2000emotion}. Actions are generated from either the model-free policy or the reflexive actions depending on whether the observation meets certain conditions such as deadly situations. For such observations, the actions proposed by the model-free policy may not align with the reflexive response, as the former generates rewarding actions while the latter is hard-wired. Therefore, future trajectories simulated using the model-free policy may be inaccurate, as they wouldn't account for the agent's reflexive responses to critical observations. This planning discrepancy could potentially be rectified by employing a distilled policy which learns from its past actions, thus gaining a comprehensive understanding of the agent's action policies including the reflexive behavior.

In order to assess our hypothesis, we investigated the performance of three types of agents: simple, shared policy, and dual policy agents, all of which demonstrate reflexive behaviors. We introduced two forms of these reflexive responses: 'flight', where the agent retreats in the opposite direction, and 'freeze', where the agent remains still when the predator comes too close. While the 'flight' response is advantageous as it aids the agent in avoiding the predator, the 'freeze' response is detrimental in our environment as it prevents the agent from escaping the predator. This can be identified from the evaluation performance of the Simple agent with 'flight' behavior (Table \ref{tab: evaluation_innate_flight}), with 'freeze' behavior (Table \ref{tab: evaluation_innate_freeze}), and without reflexive behavior from the previous section \ref{subsection: stability} (Table \ref{tab: evaluation}). We incorporated both advantageous and detrimental behaviors to postulate that the dual policy agent's performance would only improve with the detrimental reflexive behavior, given that this behavior would deviate more from the model-free policy in comparison to the advantageous behavior.

Our dual policy agent successfully reconciled the action discrepancies between the model-free policy and the reflexive behavior, particularly with the 'freeze' reflexive behavior where it significantly outperformed the shared policy agent (Figure \ref{fig: training_innate}, Table \ref{tab: evaluation_innate_freeze}). However, it didn't offer a similar performance advantage with the 'flight' reflexive behavior (Figure \ref{fig: training_innate}, Table \ref{tab: evaluation_innate_flight}). These results suggest that the dual policy agent enhances performance by learning a more comprehensive action space, which includes both model-free policy actions and non-simulatable reflexive behaviors.

\section{Discussion}
We explored a range of benefits of employing a distilled policy as a self-model for planning. First, we found that the dual policy approach with distillation stabilizes training, a finding analogous to the stability often seen employing ensemble approaches such as the double Q-learning \cite{van2016deep}. Secondly, we found that the flexible architecture of the dual policy can accelerate planning, a process that inherently requires numerous policy inferences. Moreover, our dual policy framework enhances exploration capabilities, and effectively handles sub-optimal actions that are not feasible to simulate for planning, thereby increasing robustness. These empirical findings suggest a potential parallel with the neurobiological structures of the brain, which may also harbor separate networks for similar reasons - stabilization of training, speed of planning, enhancement of exploration, and planing with holistic understanding of one's behavior. Our work thus provides insights not only in the realm of reinforcement learning, but also in understanding the principles of cognitive systems.

While our work offers promising results, it was validated within a single environment which presents a limit to the generalizability of our findings. Future work could extend this investigation to variety of environments to solidify the observed benefits and uncover any environment-specific characteristics. Moreover, there are other potential benefits of having a distilled policy as the self-model that needs to be explored. For instance, aspects such as temporal abstraction, a key concept in hierarchical reinforcement learning, could be investigated in the context of our dual policy agent. Additionally, the idea of using the distilled policy to retrain the model-free policy when it is impaired could be examined as a means of ensuring lesion stability. This could offer a robust mechanism for recovery under unexpected disruptions. Beyond the context of single agent behavior, the dual policy paradigm could also be explored in multi-agent settings in terms of 'theory of mind'. That is, understanding how an agent's self-model could help it predict and adapt to the behaviors of other agents. In summary, our findings represent an early step in the concept of self-models and their design, highlighting the potential it holds.

\begin{ack}
This research was supported by the MOTIE (Ministry of Trade, Industry, and Energy) in Korea, under Human Resource Development Program for Industrial Innovation (Global) (P0017311) supervised by the Korea Institute for Advancement of Technology (KIAT). The manuscript was revised with the help of chatGPT. None of the experiment codes were generated from chatGPT.

\end{ack}

\medskip

\bibliography{main}
\bibliographystyle{unsrt}

\end{document}